\documentclass[11pt,a4paper]{article}
\usepackage[hyperref]{naaclhlt2019}
\usepackage{times}
\usepackage{latexsym}
\usepackage{textcomp}
\usepackage{siunitx}
\usepackage{amsmath}
\usepackage{amssymb,amsfonts,bm}
\usepackage{booktabs}
\usepackage{multirow}
\usepackage{graphicx}
\usepackage{url}
\usepackage{footnote}
\usepackage[symbol,para]{footmisc}
\usepackage{xcolor}
\usepackage{xspace}
\usepackage{float}
\usepackage{placeins}
\usepackage{algorithm2e}
\usepackage{subcaption}

\PassOptionsToPackage{hyphens}{url}\usepackage{hyperref}

\aclfinalcopy 
\def\aclpaperid{1301} 

\newcommand{\parheader}[1]{{\smallskip \noindent \bf #1.}}

\newcommand\caplbl[1]{\textbf{(#1)}}

\DeclareMathOperator*{\softmax}{softmax}

\title{Distilling Task-Specific Knowledge from BERT into\\ Simple Neural Networks} 

\author{Raphael Tang\thanks{Equal contribution. Ordering decided by coin toss.}, \hspace{0.07cm} Yao Lu\footnotemark[1], \hspace{0.07cm} Linqing Liu\footnotemark[1], \hspace{0.07cm} Lili Mou, \hspace{0.07cm} Olga Vechtomova, \and Jimmy Lin\vspace{0.1cm}\\
 University of Waterloo\\
{\tt \{r33tang, yao.lu, linqing.liu\}@uwaterloo.ca}\\ {\tt doublepower.mou@gmail.com} \quad {\tt\{ovechtom, jimmylin\}@uwaterloo.ca}}

\begin{document}
\maketitle
\begin{abstract}
In the natural language processing literature, neural networks are becoming increasingly deeper and complex.
The recent poster child of this trend is the deep language representation model, which includes BERT, ELMo, and GPT.
These developments have led to the conviction that previous-generation, shallower neural networks for language understanding are obsolete.
In this paper, however, we demonstrate that rudimentary, lightweight neural networks can still be made competitive \textit{without} architecture changes, external training data, or additional input features.
We propose to distill knowledge from BERT, a state-of-the-art language representation model, into a single-layer BiLSTM, as well as its siamese counterpart for sentence-pair tasks.
Across multiple datasets in paraphrasing, natural language inference, and sentiment classification, we achieve comparable results with ELMo, while using roughly 100 times fewer parameters and 15 times less inference time.

\end{abstract}

\section{Introduction}

 In the natural language processing (NLP) literature, the march of the neural networks has been an unending yet predictable one, with new architectures constantly surpassing previous ones in not only performance and supposed insight but also complexity and depth.
 In the midst of all this neural progress, it becomes easy to dismiss earlier, ``first-generation'' neural networks as obsolete.
 Ostensibly, this appears to be true:~\citet{peters2018deep} show that using pretrained deep word representations achieves state of the art on a variety of tasks.
 Recently, \citet{devlin2018bert} have pushed this line of work even further with bidirectional encoder representations from transformers (BERT), deeper models that greatly improve state of the art on more tasks.
 More recently, OpenAI has described GPT-2, a state-of-the-art, larger transformer model trained on even more data.\footnote{\url{https://goo.gl/Frmwqe}}
 
Such large neural networks are, however, problematic in practice. Due to the large number of parameters, BERT and GPT-2, for example, are undeployable in resource-restricted systems such as mobile devices. They may be inapplicable in real-time systems either, because of low inference-time efficiency. Furthermore, the continued slowdown of Moore's Law and Dennard scaling~\cite{han2017efficient} suggests that there exists a point in time when we must compress our models and carefully evaluate our choice of the neural architecture.

 
 In this paper, we propose a simple yet effective approach that transfers task-specific knowledge from BERT to a shallow neural architecture---in particular, a bidirectional long short-term memory network (BiLSTM). 
 Our motivation is twofold:~we question whether a simple architecture actually lacks representation power for text modeling, and we wish to study effective approaches to transfer knowledge from BERT to a BiLSTM.
 Concretely, we leverage the knowledge distillation approach~\cite{ba2014deep, hinton2015distilling}, where a larger model serves as a \textit{teacher} and a small model learns to mimic the teacher as a \textit{student}. 
 This approach is model agnostic, making knowledge transfer possible between BERT and a different neural architecture, such as a single-layer BiLSTM, in our case. 
 
 To facilitate effective knowledge transfer, however, we often require a large, unlabeled dataset. The teacher model provides the probability logits and estimated labels for these unannotated samples, and the student network learns from the teacher's outputs.
 In computer vision, unlabeled images are usually easy to obtain through augmenting the data using rotation, additive noise, and other distortions.
 However, obtaining additional, even unlabeled samples for a specific task can be difficult in NLP.
 Traditional data augmentation in NLP is typically task-specific~\cite{wang2016galactic, serban2016generating} and difficult to extend to other NLP tasks. 
 To this end, we further propose a novel, rule-based textual data augmentation approach for constructing the knowledge transfer set.
 Although our augmented samples are not fluent natural language sentences, experimental results show that our approach works surprisingly well for knowledge distillation.
 
 
We evaluate our approach on three tasks in sentence classification and sentence matching. Experiments show that our knowledge distillation procedure significantly outperforms training the original simpler network alone.
To our knowledge, we are the first to explore distilling knowledge from BERT. 
With our approach, a shallow BiLSTM-based model achieves results comparable to Embeddings from Language Models (ELMo; \citealp{peters2018deep}), but uses around 100 times fewer parameters and performs inference 15 times faster.
Therefore, our model becomes a state-of-the-art ``small'' model for neural NLP.


\section{Related Work}

In the past, researchers have developed and applied various neural architectures for NLP, including convolutional neural networks~\cite{kalchbrenner2014convolutional, kim2014convolutional}, recurrent neural networks~\cite{mikolov2010recurrent, mikolov2011extensions, graves2013generating}, and recursive neural networks~\cite{socher2010learning,socher2011parsing}.
These generic architectures can be applied to tasks like sentence classification~\cite{zhang2015character, conneau2016very} and sentence matching~\cite{wan2016deep, he2016umd}, but the model is trained only on data of a particular task. 



Recently, \citet{peters2018deep} introduce Embeddings from Language Models (ELMo), an approach for learning high-quality, deep contextualized representations using bidirectional language models. With ELMo, they achieve large improvements on six different NLP tasks.
\citet{devlin2018bert} propose Bidirectional Encoder Representations from Transformers (BERT), a new language representation model that obtains state-of-the-art results on eleven natural language processing tasks. 
Trained with massive corpora for language modeling, BERT has strong syntactic ability~\cite{goldberg2019assessing} and captures generic language features. 
A typical downstream use of BERT is to fine-tune it for the NLP task at hand.
This improves training efficiency, but for inference efficiency, these models are still considerably slower than traditional neural networks.

\parheader{Model compression}
A prominent line of work is devoted to compressing large neural networks to accelerate inference.
Early pioneering works include \citet{lecun1990optimal}, who propose a local error-based method for pruning unimportant weights.
Recently, \citet{han2015deep} propose a simple compression pipeline, achieving 40 times reduction in model size without hurting accuracy.
Unfortunately, these techniques induce irregular weight sparsity, which precludes highly optimized computation routines.
Thus, others explore pruning entire filters~\cite{li2016pruning, liu2017learning}, with some even targeting device-centric metrics, such as floating-point operations~\cite{tang2018flops} and latency~\cite{chen2018constraint}.
Still other studies examine quantizing neural networks~\cite{wu2018training}; in the extreme, \citet{courbariaux2016binarized} propose binarized networks with both binary weights and binary activations.

Unlike the aforementioned methods, the knowledge distillation approach~\cite{ba2014deep,hinton2015distilling} enables the transfer of knowledge from a large model to a smaller, ``student'' network, which is improved in the process.
The student network can use a completely different architecture, since distillation works at the output level.
This is important in our case, since our research objective is to study the representation power of shallower neural networks for language understanding, while simultaneously compressing models like BERT; thus, we follow this approach in our work.
In the NLP literature, it has previously been used in neural machine translation~\cite{kim2016sequence} and language modeling~\cite{yu2018device}.



\section{Our Approach}

First, we choose the desired teacher and student models for the knowledge distillation approach. Then, we describe our distillation procedure, which comprises two major components:~first, the addition of a logits-regression objective, and second, the construction of a transfer dataset, which augments the training set for more effective knowledge transfer.

\begin{figure}
    \centering
    \includegraphics[scale=0.3]{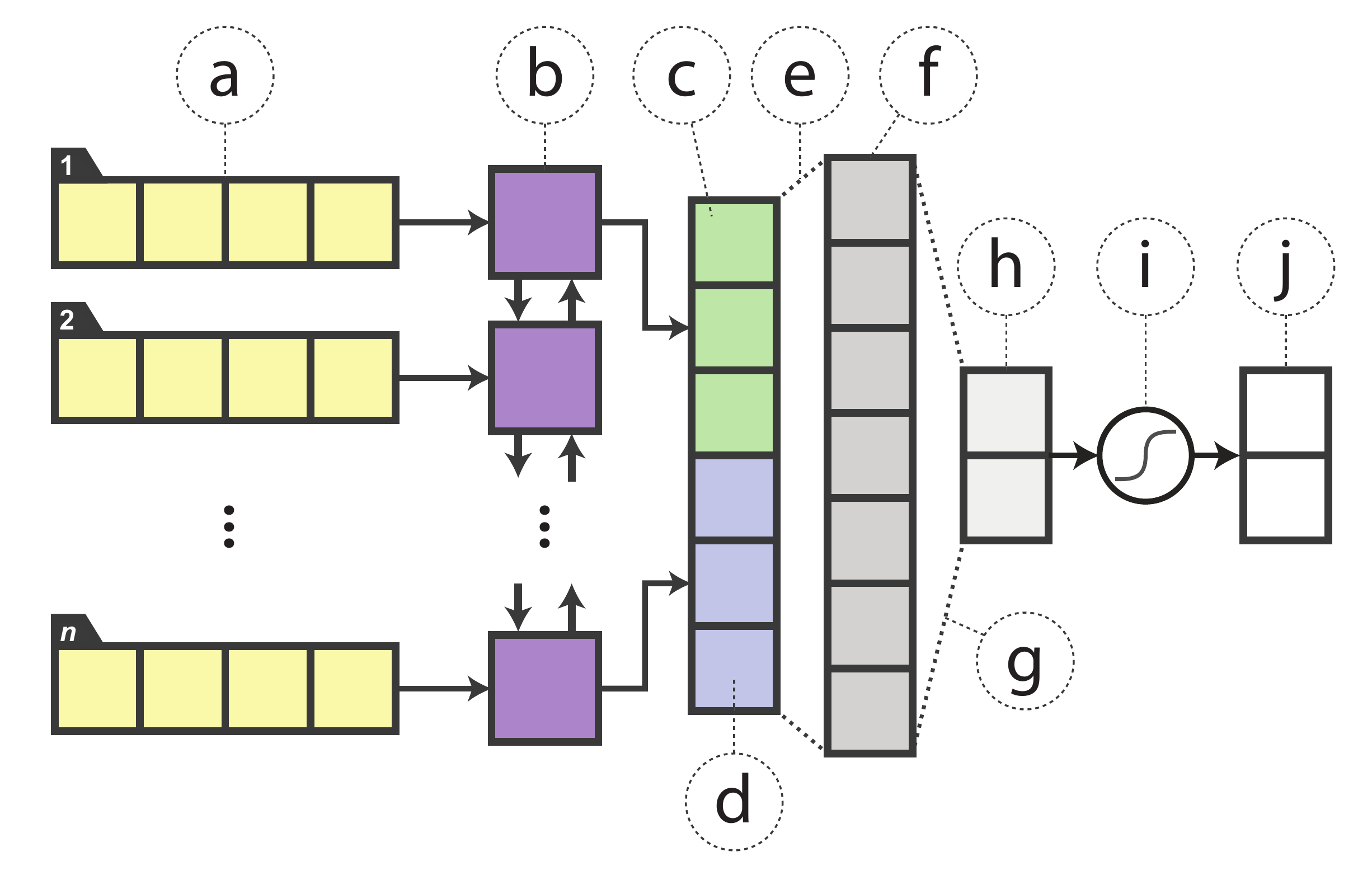}
    \caption{The BiLSTM model for single-sentence classification. The labels are \caplbl{a} input embeddings, \caplbl{b}~BiLSTM, \caplbl{c, d} backward and forward hidden states, respectively, \caplbl{e, g} fully-connected layer; \caplbl{e}~with ReLU, \caplbl{f} hidden representation, \caplbl{h} logit outputs, \caplbl{i} softmax activation, and \caplbl{j} final probabilities.}
    \label{fig:bilstm}
\end{figure}
\subsection{Model Architecture}
For the teacher network, we use the pretrained, fine-tuned BERT~\cite{devlin2018bert} model, a deep, bidirectional transformer encoder that achieves state of the art on a variety of language understanding tasks. 
From an input sentence (pair), BERT computes a feature vector $\bm h\in\mathbb{R}^{d}$, upon which we build a classifier for the task. For single-sentence classification, we directly build a softmax layer, i.e., the predicted probabilities are $\bm y^{(B)}=\softmax(W\bm h)$, where $W\in\mathbb{R}^{k\times d}$ is the softmax weight matrix and $k$ is the number of labels. For sentence-pair tasks, we concatenate the BERT features of both sentences and feed them to a softmax layer. During training, we jointly fine-tune the parameters of BERT and the classifier by maximizing the probability of the correct label, using the cross-entropy loss.



\begin{figure}
    \centering
    \includegraphics[scale=0.25]{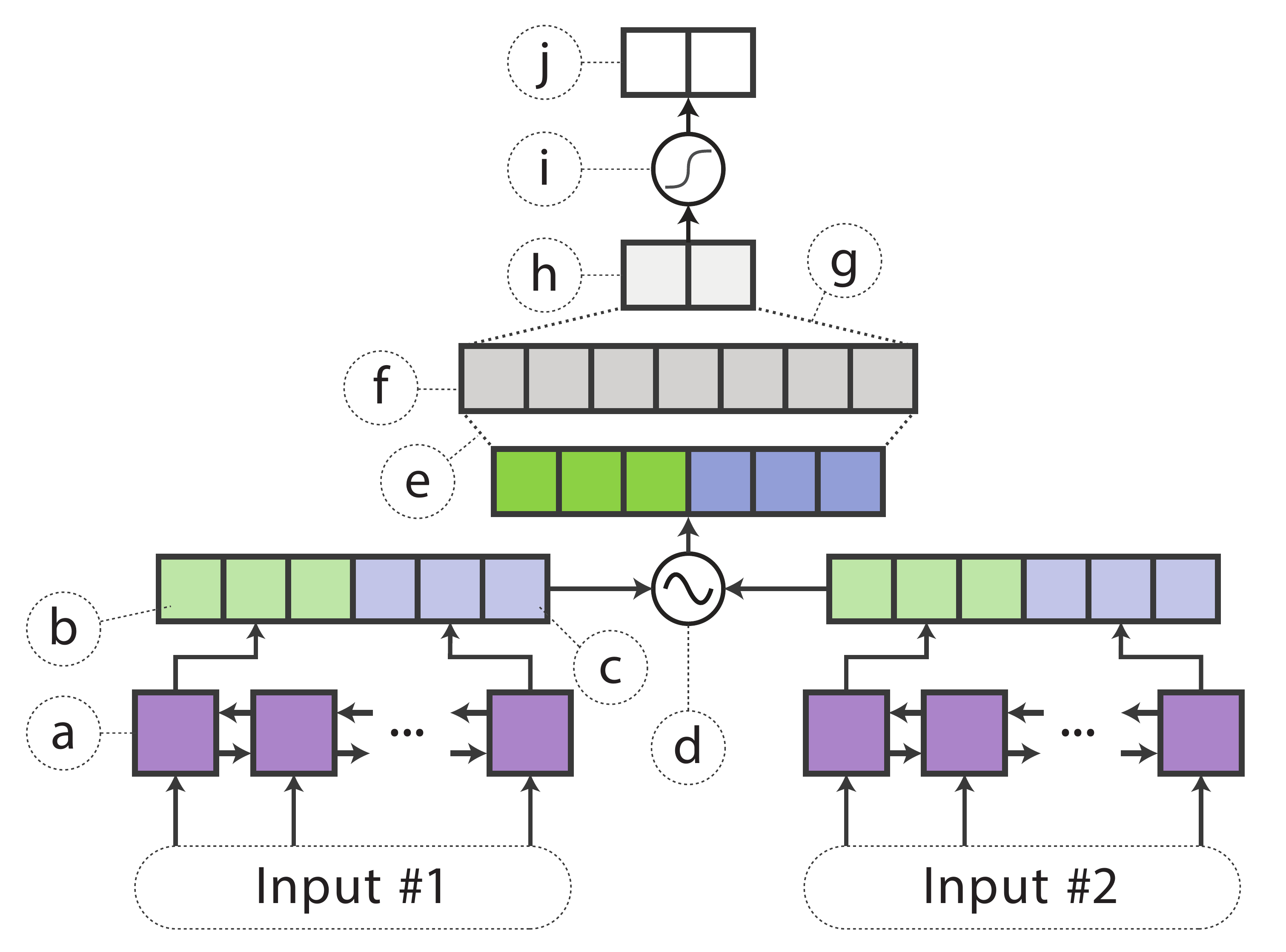}
    \caption{The siamese BiLSTM model for sentence matching, with shared encoder weights for both sentences. The labels are \caplbl{a} BiLSTM, \caplbl{b, c} final backward and forward hidden states, respectively, \caplbl{d} concatenate--compare unit, \caplbl{e, g} fully connected layer; \caplbl{e} with ReLU, \caplbl{f} hidden representation, \caplbl{h} logit outputs, \caplbl{i}~softmax activation, and \caplbl{j}~final probabilities.}
    \label{fig:bilstmsiamese}
\end{figure}

In contrast, our student model is a single-layer BiLSTM with a non-linear classifier. After feeding the input word embeddings into the BiLSTM, the hidden states of the last step in each direction are concatenated and fed to a fully connected layer with rectified linear units (ReLUs), whose output is then passed to a softmax layer for classification (Figure~\ref{fig:bilstm}). For sentence-pair tasks, we share BiLSTM encoder weights in a siamese architecture between the two sentence encoders, producing sentence vectors $\bm h_{s1}$ and $\bm h_{s2}$ (Figure~\ref{fig:bilstmsiamese}). We then apply a standard concatenate--compare operation~\cite{wang2018glue} between the two sentence vectors:~$f(\bm h_{s1}, \bm h_{s2}) = [\bm h_{s1}, \bm h_{s2}, \bm h_{s1} \odot \bm h_{s2}, |\bm h_{s1} - \bm h_{s2}|]$, where $\odot$ denotes elementwise multiplication. We feed this output to a ReLU-activated classifier.

It should be emphasized that we restrict the architecture engineering to a minimum to revisit the representation power of BiLSTM itself. We avoid any additional tricks, such as attention and layer normalization. 



\subsection{Distillation Objective}\label{sec:distillation-obj}

The distillation approach accomplishes knowledge transfer at the output level; that is, the student network learns to mimic a teacher network's behavior given any data point. In particular, \citet{ba2014deep} posit that, in addition to a one-hot predicted label, the teacher's predicted probability is also important. In binary sentiment classification, for example, some sentences have a strong sentiment polarity, whereas others appear neutral. If we use only the teacher's predicted one-hot label to train the student, we may lose valuable information about the prediction uncertainty.

The discrete probability output of a neural network is given by
\begin{equation}
    \widetilde y_i = \softmax(\bm z) = 
    \frac{\exp\{\bm w_i^\top \bm h\}}{\sum_j\exp\{\bm w_j^\top \bm h \}}
\end{equation}
where $\bm w_i$ denotes the $i$\textsuperscript{th} row of softmax weight $W$, and $\bm z$ is equivalent to $\bm w^\top \bm h$.
The argument of the softmax function is known as \textit{logits}. Training on logits makes learning easier for the student model since the relationship learned by the teacher model across all of the targets are equally emphasized \cite{ba2014deep}.



The distillation objective is to penalize the mean-squared-error (MSE) loss between the student network's logits against the teacher's logits:
\begin{align}\label{eqn:distill}
    \mathcal{L}_\text{distill} = ||\pmb z^{(B)} - \pmb z^{(S)}||^2_2
\end{align}
where $\pmb z^{(B)}$ and $\pmb z^{(S)}$ are the teacher's and student's logits, respectively. Other measures such as cross entropy with soft targets are viable as well~\cite{hinton2015distilling}; however, in our preliminary experiments, we found MSE to perform slightly better. 

At training time, the distilling objective can be used in conjunction with a traditional cross-entropy loss against a one-hot label $\bm t$, given by
\begin{align}
&\mathcal{L} =\ \alpha\cdot \mathcal{L}_\text{CE} + (1-\alpha)\cdot\mathcal{L}_\text{distill}\\ \nonumber
&=-\alpha\sum_i t_i\log y_i^{(S)} - (1-\alpha)||\pmb z^{(B)} - \pmb z^{(S)}||^2_2
\end{align}
When distilling with a labeled dataset, the one-hot target $\bm t$ is simply the ground-truth label. When distilling with an unlabeled dataset,  we use the predicted label by the teacher, i.e., $t_i=1$ if $i=\operatorname{argmax}{ \bm y^{(B)}}$ and $0$ otherwise.


\subsection{Data Augmentation for Distillation}
In the distillation approach, a small dataset may not suffice for the teacher model to fully express its knowledge~\cite{ba2014deep}. Therefore, we augment the training set with a large, unlabeled dataset, with pseudo-labels provided by the teacher, to aid in effective knowledge distillation.

Unfortunately, data augmentation in NLP is usually more difficult than in computer vision. First, there exist a large number of homologous images in computer vision tasks. CIFAR-10, for example, is a subset of the 80 million tiny images dataset~\cite{krizhevsky2009learning}. 
Second, it is possible to synthesize a near-natural image by rotating, adding noise, and other distortions, but if we manually manipulate a natural language sentence, the sentence may not be fluent, and its effect in NLP data augmentation less clear.


In our work, we propose a set of heuristics for task-agnostic data augmentation: we use the original sentences in the small dataset as blueprints, and then modify them with our heuristics, a process analogous to image distortion. Specifically, we randomly perform the following operations.

\parheader{Masking}
With probability $p_{\text{mask}}$, we randomly replace a word with \texttt{[MASK]}, which corresponds to an unknown token in our models and the masked word token in BERT. Intuitively, this rule helps to clarify the contribution of each word toward the label, e.g., the teacher network produces less confident logits for ``I \texttt{[MASK]} the comedy'' than for ``I loved the comedy.''

\parheader{POS-guided word replacement}
With probability $p_{\text{pos}}$, we replace a word with another of the same POS tag. 
To preserve the original training distribution, the new word is sampled from the unigram word distribution re-normalized by the part-of-speech (POS) tag.
This rule perturbs the semantics of each example, e.g., ``What do pigs eat?'' is different from ``How do pigs eat?''

\parheader{$\pmb n$-gram sampling}
With probability $p_{\text{ng}}$, we randomly sample an $n$-gram from the example, where $n$ is randomly selected from {$\{1, 2, \dots, 5\}$}.
This rule is conceptually equivalent to dropping out all other words in the example, which is a more aggressive form of masking.

\smallskip 

Our data augmentation procedure is as follows:~given a training example $\{w_1, \dots w_n\}$, we iterate over the words, drawing from the uniform distribution $X_i \sim \textsc{Uniform}[0, 1]$ for each $w_i$.
If $X_i < p_\text{mask}$, we apply masking to $w_i$. 
If $p_\text{mask} \leq X_i < p_\text{mask} + p_\text{pos}$, we apply POS-guided word replacement. We treat masking and POS-guided swapping as mutually exclusive: once one rule is applied, the other is disregarded.
After iterating through the words, with probability $p_\text{ng}$, we apply $n$-gram sampling to this entire synthetic example.
The final synthetic example is appended to the augmented, unlabeled dataset.

We apply this procedure $n_{\text{iter}}$ times per example to generate up to $n_\text{iter}$ samples from a single example, with any duplicates discarded. 
For sentence-pair datasets, we cycle through augmenting the first sentence only (holding the second fixed), the second sentence only (holding the first fixed), and both sentences.

\section{Experimental Setup}
For BERT, we use the large variant BERT$_\text{LARGE}$ (described below) as the teacher network, starting with the pretrained weights and following the original, task-specific fine-tuning procedure~\cite{devlin2018bert}.
We fine-tune four models using the Adam optimizer with learning rates $\{2, 3, 4, 5\}\times10^{-5}$, picking the best model on the validation set.
We avoid data augmentation during fine-tuning.

For our models, we feed the original dataset together with the synthesized examples to the task-specific, fine-tuned BERT model to obtain the predicted logits.
We denote our distilled BiLSTM trained on soft logit targets as BiLSTM$_\text{SOFT}$, which corresponds to choosing $\alpha = 0$ in Section~\ref{sec:distillation-obj}.
Preliminary experiments suggest that using only the distillation objective works best.

\subsection{Datasets}

We conduct experiments on the General Language Understanding Evaluation~(GLUE; \citealp{wang2018glue}) benchmark, a collection of six natural language understanding tasks that are classified into three categories: single-sentence tasks, similarity and paraphrase tasks, and inference tasks. Due to restrictions in time and computational resources, we choose the most widely used dataset from each category, as detailed below.

\parheader{SST-2}
Stanford Sentiment Treebank 2 (SST-2; \citealp{socher2013recursive}) comprises single sentences extracted from movie reviews for binary sentiment classification (positive vs.~negative). Following GLUE, we consider sentence-level sentiment only, ignoring the sentiment labels of phrases provided by the original dataset.

\parheader{MNLI}
The Multi-genre Natural Language Inference (MNLI; \citealp{williams2017broad}) corpus is a large-scale, crowdsourced entailment classification dataset. 
The objective is to predict the relationship between a pair of sentences as one of entailment, neutrality, or contradiction.
\mbox{MNLI-m} uses development and test sets that contain the same genres from the training set, while \mbox{MNLI-mm} represents development and test sets from the remaining, mismatched genres.

\parheader{QQP}
Quora Question Pairs (QQP; \citealp{qqp}) consists of pairs of potentially duplicate questions collected from Quora, a question-and-answer website. 
The binary label of each question pair indicates redundancy.

\subsection{Hyperparameters}
We choose either 150 or 300 hidden units for the BiLSTM, and 200 or 400 units in the ReLU-activated hidden layer, depending on the validation set performance. Following~\citet{kim2014convolutional}, we use the traditional 300-dimensional word2vec embeddings trained on Google News and multichannel embeddings. For optimization, we use AdaDelta~\cite{zeiler2012adadelta} with its default learning rate of $1.0$ and $\rho=0.95$.
For SST-2, we use a batch size of 50; for MNLI and QQP, due to their larger size, we choose 256 for the batch size. 

For our dataset augmentation hyperparameters, we fix $p_\text{mask} = p_\text{pos} = 0.1$ and $p_\text{ng} = 0.25$ across all datasets. These values have \textit{not} been tuned at all on the datasets---these are the first values we chose. We choose $n_\text{iter} = 20$ for \mbox{SST-2} and $n_\text{iter} = 10$ for both MNLI and QQP, since they are larger.

\subsection{Baseline Models}

\textbf{BERT}~\cite{devlin2018bert} is a multi-layer, bidirectional transformer encoder that comes in two variants:~BERT$_{\text{BASE}}$ and the larger BERT$_{\text{LARGE}}$. BERT$_{\text{BASE}}$ comprises 12 layers, 768 hidden units, 12 self-attention heads, and 110M parameters. BERT$_{\text{LARGE}}$ uses 24 layers, 1024 hidden units, 16 self-attention heads, and 340M parameters.

\smallskip \noindent \textbf{OpenAI GPT}~\cite{radford2018improving}
is, like BERT, a generative pretrained transformer (GPT) encoder fine-tuned on downstream tasks. 
Unlike BERT, however, GPT is unidirectional and only makes use of previous context at each time step.

\parheader{GLUE ELMo baselines} 
In the GLUE paper, \citet{wang2018glue} provide a BiLSTM-based model baseline trained on top of ELMo and jointly fine-tuned across \textit{all} tasks. This model contains 4096 units in the ELMo BiLSTM and more than 93 million total parameters. In the BERT paper, \citet{devlin2018bert} provide the same model but a result slightly different from  \citet{wang2018glue}. For fair comparison, we report both results.



\begin{table*}[t]
\centering
\scalebox{1}{
\begin{tabular}{@{}llcccc@{}}
\toprule[1pt]
\multirow{2}{*}{\textbf{\#}} & \multirow{2}{*}{Model} & \multicolumn{1}{c}{SST-2} & \multicolumn{1}{c}{QQP} & \multicolumn{1}{c}{MNLI-m}  
& \multicolumn{1}{c}{MNLI-mm} \\ \cmidrule(l){3-6} 
 & & Acc & F$_\text{1}$/Acc & Acc & Acc \\ \midrule
 
1 & BERT$_{\text{LARGE}}$~\cite{devlin2018bert} & 94.9  & 72.1/89.3 & 86.7 & 85.9 \\ 
2 & BERT$_{\text{BASE}}$~\cite{devlin2018bert} & 93.5  & 71.2/89.2 & 84.6 & 83.4 \\ 
3 & OpenAI GPT~\cite{radford2018improving} & 91.3  & 70.3/88.5 & 82.1 & 81.4 \\ 
4 & BERT ELMo baseline~\cite{devlin2018bert}  & 90.4  & 64.8/84.7 & 76.4 & 76.1 \\
5 & GLUE ELMo baseline~\cite{wang2018glue}  & 90.4  & 63.1/84.3 & 74.1 & 74.5 \\

\midrule

6 & Distilled BiLSTM$_{\text{SOFT}}$ & \textbf{90.7}  &   \textbf{68.2/88.1} & \textbf{73.0} & \textbf{72.6} \\ 


7 & BiLSTM (our implementation) & 86.7  & 63.7/86.2 & 68.7 & 68.3  \\
8 & BiLSTM (reported by GLUE) & 85.9  & 61.4/81.7 & 70.3 & 70.8\\ 
9 & BiLSTM (reported by  other papers) & 87.6$^\dagger$  & ~~~--~~~/82.6$^\ddagger$ & ~~66.9\textsuperscript{*} & ~~66.9\textsuperscript{*} \\





\bottomrule[1pt]
\end{tabular}}
\caption{Test results on different datasets. The BiLSTM results reported by other papers are drawn from \citet{zhou2016text},$^\dagger$ \citet{wang2017bilateral},$^\ddagger$ and \citet{williams2017broad}.$^*$ All of our test results are obtained from the GLUE benchmark website.}
\label{table:results}
\end{table*}

\section{Results and Discussion}

We present the results of our models as well as baselines in Table \ref{table:results}.
For QQP, we report both F$_1$ and accuracy, since the dataset is slightly unbalanced.
Following GLUE, we report the average score of each model on the datasets.

\subsection{Model Quality}

To verify the correctness of our implementation, we train the base BiLSTM model on the original labels, without using distillation (row 7).
Across all three datasets, we achieve scores comparable with BiLSTMs from previous works (rows 8 and 9), suggesting that our implementation is fair.
Note that, on MNLI, the two baselines differ by 4\% in accuracy (rows 8 and 9).
None of the non-distilled BiLSTM baselines outperform BERT's ELMo baseline (row 4)---our implementation, although attaining a higher accuracy for QQP, falls short in F$_1$ score.

We apply our distillation approach of matching logits using the augmented training dataset, and achieve an absolute improvement of 1.9--4.5 points against our base BiLSTM.
On SST-2 and QQP, we outperform the best reported ELMo model (row 4), coming close to GPT.
On MNLI, our results trail ELMo's by a few points; however, they still represent a 4.3-point improvement against our BiLSTM, and a 1.8--2.7-point increase over the previous best BiLSTM (row 8). Overall, our distilled model is competitive with two previous implementations of ELMo BiLSTMs (rows 4--5), suggesting that shallow BiLSTMs have greater representation power than previously thought.

We do not, however, outperform the deep transformer models (rows 1--3), doing 4--7 points worse, on average.
Nevertheless, our model has much fewer parameters and better efficiency, as detailed in the following section.

\subsection{Inference Efficiency}

For our inference speed and parameter analysis, we use the open-source PyTorch implementations for BERT\footnote{\url{https://goo.gl/iRPhjP}} and ELMo~\cite{Gardner2017AllenNLP}.
On a single NVIDIA V100 GPU, we perform model inference with a batch size of 512 on all 67350 sentences of the SST-2 training set.
As shown in Table~\ref{tab: speed analysis}, our single-sentence model uses 98 and 349 times fewer parameters than ELMo and BERT$_{\text{LARGE}}$, respectively, and is 15 and 434 times faster.
At 2.2 million parameters, the variant with 300-dimensional LSTM units is twice as large, though still substantially smaller than ELMo.
For sentence-pair tasks, the siamese counterpart uses no pairwise word interactions, unlike previous state of the art~\cite{he2016pairwise}; its runtime thus scales linearly with sentence length.


\begin{table}[t]
\scalebox{0.95}{
\begin{tabular}{lcc}
\toprule[1pt]
\multirow{2}{*}{} & \# of Par. & Inference Time \\
\midrule
BERT$_{\text{LARGE}}$ & 335 (349$\times$) & 1060 (434$\times$) \\ 
ELMo & 93.6 (98$\times$) & 36.71 (15$\times$) \\ 
BiLSTM$_{\text{SOFT}}$ & 0.96 (1$\times$) & 2.44 (1$\times$) \\
\bottomrule[1pt]
\end{tabular}}
\caption{Single-sentence model size and inference speed on SST-2. \# of Par. denotes number of millions of parameters, and inference time is in seconds.}
\label{tab: speed analysis}
\end{table}

\section{Conclusion and Future Work}
In this paper, we explore distilling the knowledge from BERT into a simple BiLSTM-based model.
The distilled model achieves comparable results with ELMo, while using much fewer parameters and less inference time.
Our results suggest that shallow BiLSTMs are more expressive for natural language tasks than previously thought.

One direction of future work is to explore extremely simple architectures in the extreme, such as convolutional neural networks and even support vector machines and logistic regression. Another opposite direction is to explore slightly more complicated architectures using tricks like pairwise word interaction and attention.

\section*{Acknowledgements}
This research was enabled in part by resources provided by Compute Ontario and Compute Canada. This research was also supported by the Natural Sciences and Engineering Research Council (NSERC) of Canada.



\end{document}